\documentclass[runningheads]{llncs}
\usepackage[T1]{fontenc}
\usepackage{tabu}
\usepackage{booktabs}
\usepackage{amsmath}
\usepackage{amsfonts}
\usepackage{graphicx}

\begin{document}
\newcommand\ang{70}
\newcommand\angtwo{30}
\title{LABRAD-OR: Lightweight Memory Scene Graphs for Accurate Bimodal Reasoning in Dynamic Operating Rooms}

%
\titlerunning{LABRAD-OR}

\author{Ege Özsoy, Tobias Czempiel, Felix Holm, Chantal Pellegrini, Nassir Navab}

\institute{
Computer Aided Medical Procedures, Technische Universit{\"a}t M{\"u}nchen, Germany}

\maketitle              
\begin{abstract}
Modern surgeries are performed in complex and dynamic settings, including ever-changing interactions between medical staff, patients, and equipment. The holistic modeling of the operating room~(OR) is, therefore, a challenging but essential task, with the potential to optimize the performance of surgical teams and aid in developing new surgical technologies to improve patient outcomes. The holistic representation of surgical scenes as semantic scene graphs~(SGG), where entities are represented as nodes and relations between them as edges, is a promising direction for fine-grained semantic OR understanding. We propose, for the first time, the use of temporal information for more accurate and consistent holistic OR modeling. Specifically, we introduce memory scene graphs, where the scene graphs of previous time steps act as the temporal representation guiding the current prediction. We design an end-to-end architecture that intelligently fuses the temporal information of our lightweight memory scene graphs with the visual information from point clouds and images. We evaluate our method on the 4D-OR dataset and demonstrate that integrating temporality leads to more accurate and consistent results achieving an +5\% increase and a new SOTA of 0.88 in macro F1. This work opens the path for representing the entire surgery history with memory scene graphs and improves the holistic understanding in the OR. Introducing scene graphs as memory representations can offer a valuable tool for many temporal understanding tasks.

\end{abstract}

\section{Introduction}
\begin{figure}[hbt!]
	\centering
	\includegraphics[width=1.0\linewidth]{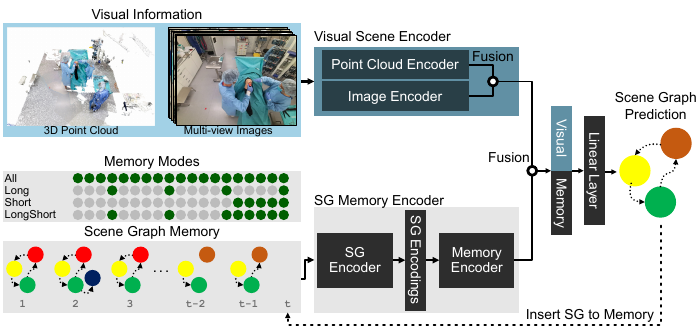}
	\caption{Overview of our bimodal scene graph generation architecture. We use the visual information extracted from point clouds and images and temporal information represented as memory scene graphs resulting in more accurate and consistent predictions.}
	\label{fig:method_overview}
\end{figure}

Surgical procedures are becoming increasingly complex, requiring intricate coordination between medical staff, patients, and equipment~\cite{maier-hein_sds_2017,lalys2014}. Effective operating room (OR) management is critical for improving patient outcomes, optimizing surgical team performance, and developing new surgical technologies~\cite{kennedymetz2020}. Scene understanding, particularly in dynamic OR environments, is a challenging task that requires holistic and semantic modeling~\cite{lalys2014,ozsoy20224d}, where both the coarse and the fine-level activities and interactions are understood. 
While many recent works addressed different aspects of this understanding, such as surgical phase recognition~\cite{twinanda2016endonet,tecno,jin2020multi}, action detection~\cite{tripnet,rendezvous}, or tool detection~\cite{jha2021kvasir,ding2022carts}, these approaches do not focus on holistic OR understanding. Most recently, Özsoy et al.~\cite{ozsoy20224d} proposed a new dataset, 4D-OR, and an approach for holistic OR modeling. They model the OR using semantic scene graphs, which summarize the entire scene at each timepoint, connecting the different entities with their semantic relationships. However, they did not propose remedies for challenges caused by occlusions and visual similarities of scenes observed at different moments of the intervention. In fact, they rely only on single timepoints for OR understanding, while temporal history is a rich information source that should be utilized for improving holistic OR modeling. 

In endoscopic video analysis, using temporality has become standard practice~\cite{twinanda2016endonet,tecno,sharma2022rendezvous,gao2021trans}. For surgical workflow recognition in the OR, the use of temporality has been explored in previous studies showcasing their effectiveness~\cite{sharghi2020automatic,jamal2022multi,mottaghi2022adaptation}. All of these methods utilize what we refer to as latent temporality, which is a non-interpretable, hidden feature representation. While some works utilize two-stage architectures, where the temporal stage uses precomputed features from a single timepoint neural network, others use 3D(2D + time) methods, directly considering multiple timepoints~\cite{czempiel2023surgical}. 

Both of these approaches have some downsides. The two-stage approaches are not trained end-to-end, potentially limiting their performance. Additionally, certain design choices must be made regarding which feature from every timepoint should be used as a temporal summary. For scene graph generation, this can be challenging, as most SGG architectures work with representations per relation and not per scene. The end-to-end 3D methods, on the other hand, are computationally expensive both in training and inference and practical hardware limitations mean they can only effectively capture short-term context. Finally, these methods can only provide limited insight into which temporal information is the most useful. 

In the computer vision community, multiple studies on scene understanding using scene graphs ~\cite{imp,3dssg} have been conducted. Ji et al.~\cite{action_genome} proposes Action Genome, a temporal scene graph dataset containing 10K videos. They demonstrate how scene graphs can be utilized to improve the action recognition performance of their model. While there have been some works~\cite{teng2021target,cong2021spatial} on using temporal visual information to enhance scene graph predictions, none consider the previous scene graph outputs as a temporal representation to enhance the future scene graph predictions.

In this paper, we propose LABRAD-OR(\textbf{L}ightweight Memory Scene Graphs for \textbf{A}ccurate \textbf{B}imodal \textbf{R}e\textbf{A}soning in \textbf{D}ynamic \textbf{O}perating \textbf{R}ooms), a novel and lightweight approach for generating accurate and consistent scene graphs using the temporal information available in OR recordings. To this end, we introduce the concept of memory scene graphs, where, for the first time, the scene graphs serve as both the output and the input, integrating temporality into the scene graph generation. Our motivation behind using scene graphs to represent memory is twofold. First, by design, they summarize the most relevant information of a scene, and second, they are lightweight and interpretable, unlike a latent feature-based representation. We design an end-to-end architecture that fuses this temporal information with visual information. This bimodal approach not only leads to significantly higher scene graph generation accuracy than the state-of-the-art but also to better inter-timepoint consistency. Additionally, by choosing lightweight architectures to encode the memory scene graphs, we can integrate the entire temporal information, as human-interpretable scene graphs, with only 40\% overhead. We show the effectiveness of our approach through experiments and ablation studies.

\section{Methodology}
In this section, we introduce our memory scene graph-based temporal modeling approach (LABRAD-OR), a novel bimodal scene graph generation architecture for holistic OR understanding, where both the visual information, as well as the temporal information in the form of memory scene graphs, are utilized. Our architecture is visualized in Fig. \ref{fig:method_overview}. 

\subsection{Single Timepoint Scene Graph Generation}
\label{single_timepoint}
We build up on the 4D-OR~\cite{ozsoy20224d} method, which uses a single timepoint for generating semantic scene graphs. The 4D-OR method extracts human and object poses and uses them to compute point cloud features for all object pairs. Additionally, image features are incorporated into the embedding to improve the recognition of details. These representations are then further processed to generate object and relation classes and are fused to generate a comprehensive scene graph.

\subsection{Scene Graphs as Memory Representations}
\label{memory_sg}
In this study, we investigate the potential of using scene graphs from previous timepoints, which we refer to as "memory scene graphs", to inform the current prediction. Unlike previous research that treated scene graphs only as the final output, we use them both as input and output. Scene graphs are particularly well-suited for encoding scene information, as they are low-dimensional and interpretable while capturing and summarizing complex semantics. To create a memory representation at a timepoint T, we use the predicted scene graphs from timepoints 0 to T-1 and employ a neural network to compute a feature representation. This generic approach allows us to easily fuse the scene graph memory with other modalities, such as images or point clouds.

\noindent \textbf{Memory Modes:} While our efficient memory representation allows us to look at all the previous timesteps, this formulation has two downsides. Surgical duration differs between procedures, and despite the efficiency of scene graphs, prolonged interventions can still be costly. Second, empirically we find that seeing the entire memory leads to prolonged training time and can cause overfitting. To address this, we propose four different memory modes, "All", "Short", "Long", and "LongShort". The entire surgical history is visible only in the "All" mode. In the "Short" mode, only the previous $S$ scene graphs are utilized, while in the "Long" mode, every $S.th$ scene graph is selected using striding. In "LongShort" mode, both "Long" and "Short" modes are combined. The reasoning behind this approach is that short-term context should be observed in detail, while long-term context can be viewed more sparsely. This reduces computational overhead compared to "All" and leads to better results with less overfitting, as observed empirically. The value of $S$ is highly dependent on the dataset and the surgical procedures under analysis.

\begin{figure}[hbt!]
	\centering
	\includegraphics[width=1.0\linewidth]{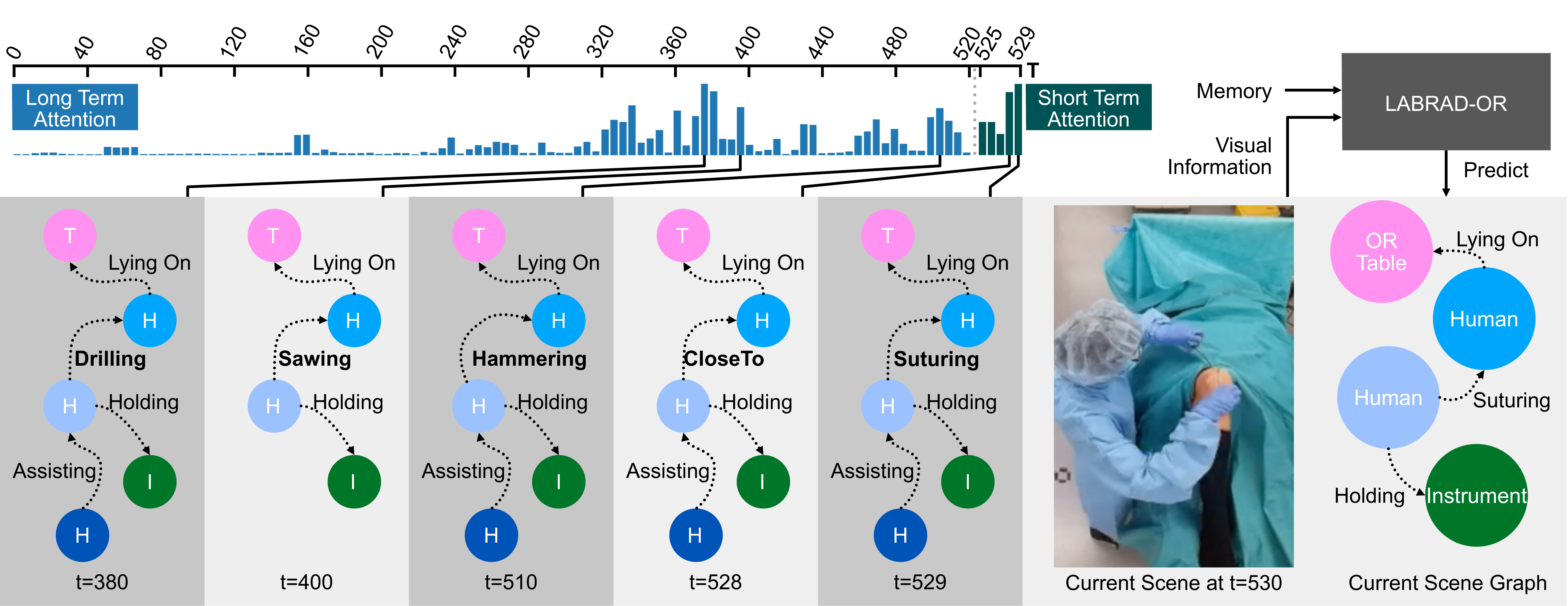}
	\caption{Visualization of both the "Short" and "Long" memory attention while predicting for the timepoint t=530. While "Short" attends naturally to the nearest scene graphs, "Long" seems to be concentrating on previous key moments of the surgery, such as "drilling", "sawing", "and hammering". The graphs are simplified for clarity. The shown "current scene graph" is the correct prediction of our model.}
	\label{fig:attention_visualization}
\end{figure}

\subsection{Architecture Design}
We extract the visual information from point clouds and, optionally, images using a visual scene encoder, as described in \ref{single_timepoint}. To integrate the temporal information, we convert each memory scene graph into a feature vector using a graph neural network. Then we use a Transformer block~\cite{transformers} to summarize the feature vectors from the $\#T$ memory scene graphs into a single feature vector, which we refer to as the memory representation. Finally, this representation is concatenated with the visual information, forming a bimodal representation. Intuitively, this allows our architecture to consider both the long-term history, such as previous key surgical steps and the short-term history, such as what was just happening. 

\noindent \textbf{Memory Augmentations:}
While we use the predicted scene graphs from previous timepoints for inference, as these are not available during training, we use the ground truth scene graphs for training. However, training with ground truth scene graphs and evaluating with predicted scene graphs can decrease test performance, as the predicted scene graphs are imperfect. To increase our robustness towards this, we utilize memory augmentations during training. Concretely, we randomly replace part of either the short-term memory (timepoints closest to the timepoint of interest) or the long-term memory (timepoints further away from the timepoint of interest) with a special "UNKNOWN" token. Intuitively, this forces our model to rely on the remaining information and better deal with wrong predictions in the memory during inference. 

\noindent \textbf{Timepoint of Interest(ToI) Positional Ids:}
Transformers~\cite{transformers} employ positional ids to encode the \textit{absolute location} of each feature in a sequence. However, as we are more interested in the relative distance of the features, we introduce Timepoint of Interest(ToI) positional ids to encode the distance of every feature to the current timepoint $T$. This allows our network to assign meaning to the relative distance to other timepoint features rather than their \textit{absolute locations}.

\noindent \textbf{Multitask Learning:}
In the scene graph generation task, the visual and temporal information are used together. In practice, we found it valuable to introduce a secondary task, which can be solved only using temporal information. We propose the task of "main action recognition", where instead of the scene graph, only the interaction of the head surgeon to the patient is predicted, such as "sawing" or "drilling". During training, in addition to fusing the memory representation with the visual information,  we use a fully connected layer to estimate the main action from the memory representation directly. Learning both scene graph generation and main action recognition tasks simultaneously gives a more direct signal to the memory encoder, resulting in faster training and improved performance. 

\section{Experiments}
\noindent \textbf{Dataset}: We use the 4D-OR~\cite{ozsoy20224d} dataset following the official train, validation, and test splits. It comprises ten simulated knee surgeries recorded using six Kinect cameras at 1 fps. Both the 3D point cloud, as well as multiview images are provided for all 6734 scenes. Each scene additionally includes a semantic scene graph label, as well as clinical role labels for staff.

\noindent \textbf{Model Training}: Our architecture consists of two components implemented in PyTorch 1.10, a visual model and a memory model. For our visual model, we use the current SOTA model from 4D-OR~\cite{ozsoy20224d}, which uses Pointnet++~\cite{pointnet++} as the point cloud encoder, and EfficientNet-B5~\cite{efficientnet} as the image encoder. We could improve the original 4D-OR results through longer training than in the original code. As our memory model, we use a combination of Graphormer~\cite{graphormer}, to extract features from individual scene graphs and Transformers~\cite{transformers}, to fuse the features into one memory representation. The visual scene encoder is initialized in all our experiments with the best-performing visual-only model weights. We use the provided human and object pose predictions from 4D-OR and stick to their training setup and evaluation metrics. We use memory augmentations, timepoint of interest positional ids, end-to-end training, and multitask learning. The memory encoders are purposefully designed to be lightweight and fast. Therefore, we use a hidden dimension of 80 and only two layers. We use $S$, to control both the stride of the "Long" mode and the window size of the "Short" mode and set it to 5. The choice of $S$ ensures we do not miss any phase, as all phases last longer than 5 timepoints while reducing the computational cost. Unless otherwise specified, we use "LongShort" as memory mode and train all models until the validation performance converges.\looseness=-1

\noindent \textbf{Evaluation Metrics}: We use the official evaluation metrics from 4D-OR for semantic scene graph generation and the role prediction downstream tasks. In both cases, a macro F1 over all the classes is computed. Further, we introduce a consistency metric, where first, for each timepoint, a set of predicates $P_t$, such as \{"assisting", "drilling", "cleaning"\} is extracted from the scene graphs. Then, for two timepoints $T$ and $T$-$1$, the intersection of union(IoU) between $P_t$ and $P_{t-1}$ is computed. This is repeated for all pairs of adjacent timepoints in a sequence, and the IoU score is averaged over them to calculate the consistency score. 

\begin{table}[t]
    \centering
    \caption{We compare our results to the current SOTA, 4D-OR, on the test set. We experimented with different hyperparameters and found that longer training can improve the 4D-OR results. We report both the original 4D-OR, and the longer trained results, indicated by $\dagger$, and a latent-based temporality(LBT) baseline, and compare LABRAD-OR to them. All methods use both point clouds and images as visual input.}
    \begin{tabu} to 0.9\textwidth { X[c] | X[c] X[c] X[c] X[c]}
    \toprule
     Method & 4D-OR~\cite{ozsoy20224d} & 4D-OR$\dagger$~\cite{ozsoy20224d} & LBT & LABRAD-OR \\ 
        \cmidrule(){1-5}
		 Macro F1 & 0.75 & 0.83 & 0.86 & \textbf{0.88}  \\
   \bottomrule
   \end{tabu}
    \label{tab:sota_comparison}
\end{table}

\begin{table}[t]
    \centering
    \caption{We demonstrate the impact of temporal information on the consistency of our results. We compare only using the point cloud(PC), using images(Img) in addition to point clouds, and temporality(T). We also show the ground truth(GT) consistency score, which should be considered the ceiling for all methods.}
    \begin{tabu} to 0.95\textwidth { X[c] | X[c] X[c] X[c] X[c] | X[c]}
    \toprule
     Method & PC & PC+Img & PC+T & PC+Img+T & GT \\ 
        \cmidrule(){1-6}
		 Consistency & 0.83 & 0.84 & 0.86 & \textbf{0.87} & 0.9 \\
    \bottomrule
   \end{tabu}
    \label{tab:consistency_comparison}
\end{table}

\begin{figure}[hbt!]
	\centering
	\includegraphics[width=1.0\linewidth]{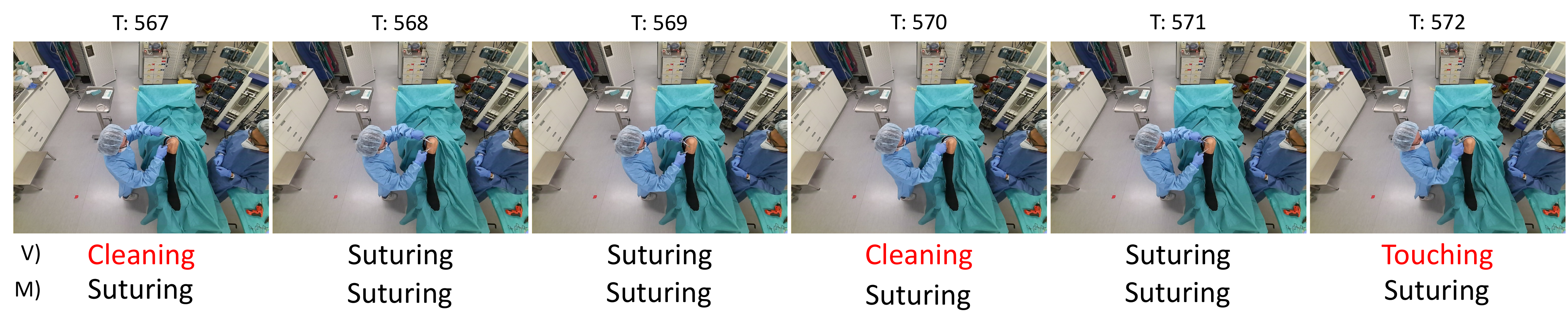}
	\caption{Qualitative example on the improvement of the scene graph consistency. For clarity, only the "main action" is shown, while only relying on the visual information (V) compared to also using our proposed SG memory (M).}
	\label{fig:consistency_qualitative}
\end{figure}

\begin{table}[t]
\begin{minipage}[]{.45\textwidth}
    \caption{We do an ablation study on the impact of memory augmentations, Timepoint of interest~(ToI) positional ids, end-to-end training, and multitask learning by individually disabling them. The experiments use the "All" memory mode and take only point cloud as visual input.}
    \begin{tabu} to 1.0\textwidth { X[c] | X[c]}
    \toprule
     Method & F1  \\ 
        \cmidrule(){1-2}
          All Techniques &  \textbf{0.86} \\
		w/o Memory Aug. & 0.82 \\
		 w/o ToI Pos Ids &  0.83  \\
		 w/o E2E & 0.84 \\
		 w/o Multitask &  0.85 \\
   \bottomrule
   \end{tabu}
    \label{tab:technique_ablations}
\end{minipage}\hfill
\begin{minipage}[]{.45\textwidth}
    \caption{Ablation study on using different memory modes, affecting which temporal information is seen. All experiments use memory augmentations, Timepoint of Interest~(ToI) positional ids, end-to-end training and multitask learning and only take point clouds as visual input.}
    \begin{tabu} to 1.0\textwidth {X[c] X[c] X[c] | X[c]}
    \toprule
     All & Short & Long & F1  \\ 
        \cmidrule(){1-4}
    \checkmark  &   & &  0.86 \\
    & \checkmark &  & 0.85 \\
    &  & \checkmark &  \textbf{0.87}  \\
    & \checkmark  & \checkmark & \textbf{0.87} \\
    \bottomrule
      & ~          \\
   \end{tabu}
    \label{tab:memory_ablation}
\end{minipage}
\end{table}

\begin{table}[t]
    \centering
    \caption{Comparison of LABRAD-OR and 4D-OR on the downstream task of role prediction.}
    \begin{tabu} to 0.6\textwidth { X[c] | X[c] X[c] }
    \toprule
     Method & 4D-OR~\cite{ozsoy20224d} & LABRAD-OR \\ 
        \cmidrule(){1-3}
		 Macro F1 & 0.85 & \textbf{0.89}  \\
   \bottomrule
   \end{tabu}
    \label{tab:role_prediction}
\end{table}

\section{Results and Discussion}
\noindent\textbf{Scene Graph Generation}
In Tab. \ref{tab:sota_comparison}, we compare our best-performing model LABRAD-OR against the previous SOTA on 4D-OR. We build a latent-based temporal baseline~(LBT), which uses a mean of the pairwise relation features as representation per timepoint, which then gets processed analogously with a Transformer architecture. Overall, LABRAD-OR increases the F1 results for all predicates, significantly increasing SOTA from 0.75 F1 to 0.88 F1. We also show that LABRAD-OR improves the F1 score compared to both the longer trained 4D-OR and LBT by 5\% and 2\%, respectively, demonstrating both the value of temporality for holistic OR modeling as well as the effectiveness of memory scene graphs. Additionally, in Tab. \ref{tab:consistency_comparison}, we show the consistency improvements achieved by using temporal information, from 0.84 to 0.87. Notably, a perfect consistency score is not 1.0, as the scene graph naturally changes over the surgery. Considering the ground truth consistency is at 0.9, LABRADOR-OR (0.87) exhibits superior consistency compared to the baselines without being excessively smooth. A qualitative example of the improvement can be seen in Fig. \ref{fig:consistency_qualitative}. While the visual-only model confuses "suturing" for "cleaning" or "touching", our LABRAD-OR model with memory scene graphs identifies all correctly "suturing". 

\noindent\textbf{Clinical Role Prediction}
We also compare LABRAD-OR to 4D-OR on the downstream task of role prediction in Tab. \ref{tab:role_prediction}, where the only difference between the two is the improved predicted scene graphs used as input for the downstream task. We improve the results by 4\%, showing our improvements in scene graph generation also translate to downstream improvements. 

\noindent\textbf{Ablation Studies}
We conduct multiple ablation studies to motivate our design choices. In Tab. \ref{tab:technique_ablations}, we demonstrate the effectiveness of the different components of our method and see that both memory augmentations and ToI positional ids are crucial and significantly contribute to the performance, whereas E2E and multitask have less but still measurable impact. We note that multitask learning also helps in stabilizing the training. In Tab. \ref{tab:memory_ablation}, we ablate the different memory modes, "Short", "Long", "LongShort", and "All", and find that the strided "Long" mode is the most important. Where the "Short" mode can often lead to insufficient memory, "All" can lead to overfitting. Both the "Long" and "LongShort" perform similarly and are more efficient than "All". We use "LongShort" as our default architecture to guarantee no short-term history is overseen.
Comparing the "LongShort" F1 result of 0.87 when only using point clouds as visual input to 0.88 when using both point cloud and images, it can be seen that images still provide valuable information, but significantly less than for Özsoy et al.~\cite{ozsoy20224d}.

\section{Conclusion}
We propose LABRAD-OR, a novel lightweight approach based on human interpretable memory scene graphs. Our approach utilizes both the visual information from the current timepoint and the temporal information from the previous timepoints for accurate bimodal reasoning in dynamic operating rooms. Through experiments, we show that this leads to significantly improved accuracy and consistency in the predicted scene graphs, consistency, and an increased score in the downstream task of role prediction. We believe LABRAD-OR offers the community an effective and efficient way of using temporal information for a holistic understanding of surgeries.\

\section{Acknowledgements}
This work has been partially supported by Stryker. Authors would like to thank Carl Zeiss AG for their support of Felix Holm.

%
%
\bibliographystyle{splncs04}
\bibliography{main}

\end{document}